\documentclass[iicol]{sn-jnl}

\usepackage{graphicx}%
\usepackage{multirow}%
\usepackage{amsmath,amssymb,amsfonts}%
\usepackage{amsthm}%
\usepackage{mathrsfs}%
\usepackage[title]{appendix}%
\usepackage{xcolor}%
\usepackage{textcomp}%
\usepackage{manyfoot}%
\usepackage{booktabs}%
\usepackage{algorithm}%
\usepackage{algorithmicx}%
\usepackage{algpseudocode}%
\usepackage{listings}%

\usepackage{caption}
\usepackage{subcaption}
\usepackage[export]{adjustbox}

\usepackage{multirow}
\usepackage{multicol}
\usepackage{bbding}

\newcommand{\ie}{\textit{i.e.} }
\newcommand{\eg}{\textit{e.g.,} }
\newcommand{\vs}{\textit{vs.} }

\newcommand{\fig}{Figure}
\newcommand{\tab}{Table}
\newcommand{\eqn}{Equation}
\newcommand{\alg}{Algorithm}

\begin{document}

\title[Article Title]{Continual Learning with Strong Experience Replay}


\author*[1]{\fnm{Tao} \sur{Zhuo}}\email{zhuotao724@gmail.com}

\author[1]{\fnm{Zhiyong} \sur{Cheng}}\email{jason.zy.cheng@gmail.com}

\author[1]{\fnm{Zan} \sur{Gao}}\email{zangaosh4522@gmail.com}

\author[2]{\fnm{Hehe} \sur{Fan}}\email{hehe.fan.cs@gmail.com}

\author[3]{\fnm{Mohan} \sur{Kankanhalli}}\email{mohan@comp.nus.edu.sg}

\affil[1]{\orgdiv{Shandong Artificial Intelligence Institute},  \orgname{Qilu University of Technology (Shandong Academy of
Sciences}, \orgaddress{\city{Jinan}, \country{China}}}

\affil[2]{\orgdiv{College of Computer Science and Technology}, \orgname{Zhejiang University}, \orgaddress{\city{HangZhou}, \country{China}}}

\affil[3]{\orgdiv{School of Computing}, \orgname{National University of Singapore}, \orgaddress{\country{Singapore}}}





\abstract{Continual Learning (CL) aims at incrementally learning new tasks without forgetting the knowledge acquired from old ones. Experience Replay (ER) is a simple and effective rehearsal-based strategy, that optimizes the model with current training data and a subset of old samples stored in a memory buffer. Although various ER extensions have been developed in recent years, the updated model suffers from the risk of overfitting the memory buffer when a few previous samples are available, leading to forgetting. In this work, we propose a Strong Experience Replay (SER) method that utilizes two consistency losses between the new model and the old one to further reduce forgetting. Besides distilling past experiences from the data stored in the memory buffer for backward consistency, we additionally explore future experiences of the old model mimicked on the current training data for forward consistency. Compared to previous methods, SER effectively improves the model generalization on previous tasks and preserves the learned knowledge. Experimental results on multiple image classification datasets show that our SER method surpasses the state-of-the-art methods by a noticeable margin. Our code is available at: {\color{magenta}https://github.com/visiontao/cl}.}




\maketitle

\section{Introduction}
Deep neural networks have been widely used in computer vision tasks, such as object detection in images and action recognition in videos. Conventional deep neural networks are usually trained offline with the assumption that all data is available. However, for a stream of nonstationary tasks, a deep model has to continually learn new tasks without forgetting the acquired knowledge from old ones~\cite{PANS2017_kirkpatrick, CS1995_Robins}. Due to the previously seen samples being unavailable for joint training, simply finetuning the model usually leads to drastic performance degradation on old tasks. This phenomenon is known as catastrophic forgetting and it seriously limits the applications in practice. To address this issue, Continual Learning (CL)~\cite{TPAMI2017_Li, NIPS2020_Buzzega} aims at preserving the acquired knowledge when learning new tasks in dynamic environments.

\begin{figure}[!t]
\centering
\includegraphics[width=0.4\textwidth]{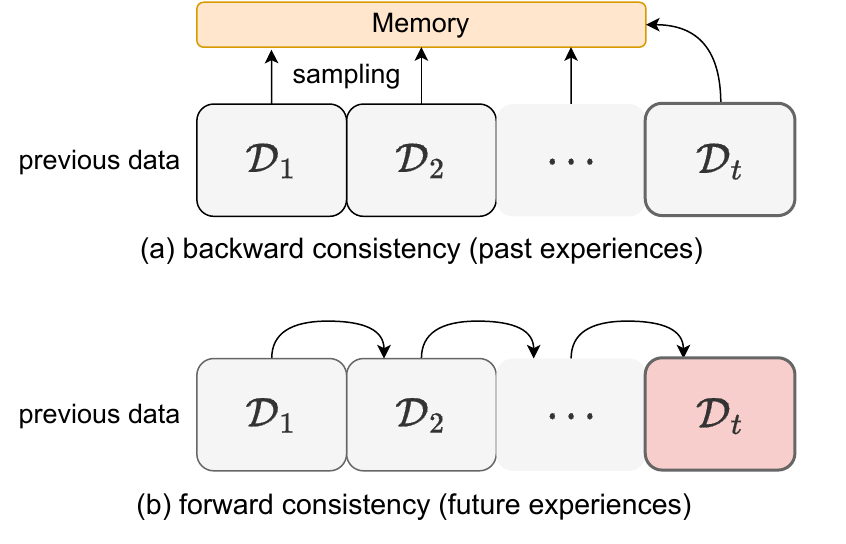}
\caption{Illustration of the backward and forward consistency. The main difference is using different data to seek consistent predictions between models. (1) The backward consistency memorizes past experiences by employing a small subset of the previous data stored in the memory buffer. (2) The forward consistency distills future experiences of the old model by leveraging all the current training data. Besides, it is able to propagate consistent predictions to new tasks over time.}
\label{fig_consistency}
\end{figure}

The core challenge in CL is to strike an optimal balance between plasticity and the stability of the model~\cite{CVPR2023_Kim}. Without revisiting any previously seen samples, it is hard to guarantee that a trained deep model with updated parameters still well fits the data of previous tasks. Therefore, the rehearsal-based technique~\cite{NIPS2020_Buzzega, ECCV2020_Hayes, ECCV2020_Iscen, ICCV2021_Verwimp, CVPR2021_Bang, CVPR2022_Wang} reduces the difficulty of CL by jointly training the model on the new data and a subset of previous data stored in the memory buffer. To leverage the memory buffer efficiently, Experience Replay (ER)~\cite{ICLR2019_Riemer} employs a reservoir sampling method to update the memory buffer over time. Then each data has the same probability to be stored in the memory buffer without knowing the length of the input data stream in advance. Based on such a simple strategy, ER effectively retains the acquired knowledge in various scenarios.

In order to further reduce the number of samples stored in the memory buffer and further mitigate forgetting, recent CL methods~\cite{NIPS2020_Buzzega, ICLR2022_AraniSZ, AAAI2023_Sarfraz} extend ER with different strategies. For example, MIR~\cite{NIPS2019_Aljundi} retrieves interfered samples stored in the memory buffer instead of random sampling. MER~\cite{ICLR2019_Riemer} combines experience replay with meta-learning to maximize transfer and minimize interference based on future gradients. DRI~\cite{AAAI2022_Wang} produces imaginary data and leverages knowledge distillation~\cite{Arxiv2015_Hinton} for retrieving past experiences in a balanced manner. Additionally, to prevent drastic model changes, rehearsal-based methods often incorporate a consistency loss for the model training. DER++~\cite{NIPS2020_Buzzega} mixes rehearsal with knowledge distillation in the memory buffer. CLS-ER~\cite{ICLR2022_AraniSZ} adopts a dual-memory method to maintain short-term and long-term semantic memories, and it also incorporates a consistency loss to prevent drastic model changes. TAMiL~\cite{ICLR2023_Bhat} entails both experience rehearsal and self-regulated scalable neurogenesis to further mitigate catastrophic forgetting. However, these methods suffer from the risk of overfitting the memory buffer when a few previous samples are available~\cite{ICCV2021_Verwimp}, leading to forgetting.

In this work, we propose a Strong Experience Replay (SER) method that extends ER with two consistency losses. Besides memorizing the label of previous data, we seek consistent prediction distributions between the new model and the old one. Ideally, when a new model has the same output logits as its original ones, the acquired knowledge can be regarded as well preserved. However, due to the samples of new classes being unseen to the old model, this goal cannot be achieved in practice. Instead, we expect the output logits of an updated model can approximate its original ones by using a consistency loss. Different from previous methods~\cite{NIPS2020_Buzzega, CVPR2022_Wang_vit, ICLR2022_AraniSZ} that seek consistency on the data stored in a limited memory buffer only, we additionally seek consistent predictions on the current training data, improving the model generalization on previous tasks and further reducing forgetting.

As illustrated in \fig~\ref{fig_consistency}, the backward consistency memorizes past experiences from the previous samples stored in the memory buffer. However, when a limited number of previous samples are available, using backward consistency may lead to a local optimum solution, resulting in overfitting. To alleviate this issue, we develop a forward consistency loss to improve the model generalization on previous tasks. Specifically, by mimicking the future experiences of the old model on the current training data, seeking consistent predictions will increase the overlapping predictions~\cite{ICML2020_Knoblauch} between the new model and the old one. Besides, since the parameters of the old model are frozen, the previous knowledge can be well preserved when optimizing the model on new tasks. Compared to backward consistency, forward consistency leverages the complete training data and it can propagate prediction distributions to new tasks over time. 

The proposed SER method effectively reduces the forgetting issue by combining both backward and forward consistency. Compared to the closest method DER++~\cite{NIPS2020_Buzzega}, the SER method additionally incorporates a forward consistency loss for model training. Despite its simplicity, extensive experiments on multiple image classification benchmark datasets show that SER outperforms DER++ by a large margin in class-incremental scenarios, \eg 9.2\% on the CIFAR100 dataset (20 tasks) and more than 17.54\% on the TinyImageNet dataset (10 tasks), see \tab~\ref{tbl_cifar100} and \ref{tbl_all}. Furthermore, comprehensive experiments on multiple image classification datasets also demonstrate the superiority of our SER method surpasses the state-of-the-art methods by a noticeable margin, especially when a few previous samples are available.

In summary, our contributions are as follows:
\begin{enumerate}
    \item We propose a forward consistency loss that improves model generalization by seeking consistent prediction distributions on the current training data. Compared to the backward consistency loss used in previous approaches, the forward consistency loss can efficiently leverage the complete training data and propagate consistent prediction distributions to new tasks over time.
    
    \item We propose a Strong Experience Replay (SER) method that distills the acquired knowledge of a trained model from both past experiences stored in the memory buffer and future experiences mimicked on the current training data. Despite its simplicity, SER greatly reduces the forgetting issue.
    
    \item The proposed method is simple and effective, extensive experiments on multiple image classification datasets show that SER outperforms the state-of-the-art methods by a large margin when a few previous samples are available.
\end{enumerate}

\section{Related Work}

\subsection{Rehearsal-based Methods}
Rehearsal-based methods~\cite{CVPR2017_Rebuffi, NIPS2019_Rolnick, CVPR2021_Bang, AAAI2022_Wang, ICLR2022_Yoon} reduce catastrophic forgetting by replaying a subset of previously seen samples stored in a memory buffer. Experience Replay (ER)~\cite{ICLR2019_Riemer} is a simple but effective rehearsal-based method that jointly trains the model with current data and a mini-batch of randomly selected old samples. Besides, it applies a reservoir sampling strategy to update the memory buffer over time. Based on the core idea of ER, recent CL approaches further reduce forgetting with various techniques. For example, GSS~\cite{NIPS2019_Aljundi} stores optimally chosen examples in the memory buffer. GEM~\cite{NIPS2017_Lopez} and AGEM~\cite{ICLR2019_chaudhry} leverage episodic memory to avoid forgetting and favor positive backward transfer. ERT~\cite{ICPR2021_Buzzega} adopts a balanced sampling method and bias control. MER~\cite{ICLR2019_Riemer} considers experience replay as a meta-learning problem to maximize transfer and minimize interference. iCaRL~\cite{CVPR2017_Rebuffi} uses a nearest-mean-of-exemplars classifier and an additional memory buffer for model training. Besides, it adopts a knowledge distillation method to reduce forgetting further. RM~\cite{CVPR2021_Bang} develops a novel memory management strategy based on per-sample classification uncertainty and data augmentation to enhance the sample diversity in the memory. DER++~\cite{NIPS2020_Buzzega} mixes rehearsal with knowledge distillation and regularization. CLS-ER~\cite{ICLR2022_AraniSZ} adopts a dual-memory experience replay method to maintain short-term and long-term semantic memories. LVT~\cite{CVPR2022_Wang_vit} designs a vision transformer for continual learning with replay. SCoMMER~\cite{AAAI2023_Sarfraz} enforces activation sparsity along with a complementary semantic dropout mechanism to encourage consistency. Different from these methods that distill past experiences from a limited memory buffer only, we also explore future experiences mimicked on the current training data, which improves the model generalization on previous tasks.

\subsection{Regularization-based Methods}
Regularization-based methods usually incorporate an additional penalty term into the loss function to prevent model changes in parameter or prediction spaces~\cite{TIP2022_Ji}. Elastic Weight Consolidation (EWC)~\cite{PANS2017_kirkpatrick, ICML2018_Schwarz}, Synaptic Intelligence (SI)~\cite{ICML2017_Zenke}, and Riemmanian Walk (RW)~\cite{ECCV2018_Chaudhry} prevent the parameter changes between the new model and the old one. LwF~\cite{TPAMI2017_Li} and PASS~\cite{CVPR2021_Zhu} mitigate forgetting with task-specific knowledge distillation. Without any previous data for replay, recent studies~\cite{NIPS2020_Buzzega, NeuComp2021} show that these methods usually fail to handle class-incremental scenarios. Therefore, the regularization-based strategy is often simultaneously used with other CL approaches for better performance, such as iCaRL~\cite{CVPR2017_Rebuffi} and DER++~\cite{NIPS2020_Buzzega, TPAMI2022_Boschini}. In this work, besides memorizing ground truth labels of previous samples, we also seek consistent logit distributions. Therefore, we incorporate two consistency losses as model regularization to reduce forgetting.

\subsection{Other CL Methods}
The early CL methods usually adopt a task identity to learn task-specific knowledge. For example, PNN (Progressive Neural Networks)~\cite{Arxiv2016_Rusu} instantiates new networks incrementally and adopts a task identity at both training and inference times. However, identity might be unknown in inference time, limiting its applications. Recently, DER~\cite{CVPR2021_Yan} expanded a new backbone per incremental task without using task identities during the testing time. FOSTER~\cite{ECCV2022_FOSTER} adds an extra model compression stage to maintain limited model storage. L2P and Dualprompt~\cite{CVPR2022_Wang, ECCV2022_Wang} adopt a prompt pool to learn and extract task-specific knowledge. In addition, for rehearsal-based methods, storing raw samples is impossible sometimes due to privacy concerns, some generative approaches~\cite{NIPS2017_Shin, ICCV2021_Smith} attempt to use a data generator (such as GAN~\cite{NIPS2014_Goodfellow}) to produce synthetic data for replay. But it needs a long time for model training and the generated data might be unsatisfactory for complex datasets. Additionally, to distill more knowledge from old tasks, DMC~\cite{WACV2020_Zhang} and GD~\cite{ICCV2019_Lee} sample external data to assist model training. Although much progress has been achieved in recent years, efficient CL with less forgetting remains a challenging problem.

\section{Method}
In this section, we first introduce the problem formulation of continual learning and the conventional experience replay method. Then we describe the proposed Strong Experience Replay (SER) method in detail.

\subsection{Problem Formulation}
Formally, given a sequence of $T$ nonstationary datasets $\{\mathcal{D}_1, \cdots, \mathcal{D}_T\}$, $(x, y) \in \mathcal{D}_t$ represents the samples $x$ with corresponding ground truth labels $y$ of the $t$-th task. At a time step, $t \in \{1, \cdots, T\}$, the goal of a CL method is to sequentially update the model on $\mathcal{D}_t$ without forgetting the previously learned knowledge on $\{\mathcal{D}_1, \cdots, \mathcal{D}_{t-1}\}$. 

In general, when all training data is available in advance, the loss function of conventional learning can be formulated as:
\begin{equation}
    \mathcal{L} = \sum_{t=1}^T \mathbb{E}_{(x, y) \sim \mathcal{D}_t}[\ell_{ce}(f(x; \theta), y)],
\end{equation}
where $\ell_{ce}$ denotes cross-entropy loss and $f(x; \theta)$ denotes the network's output logits, $\theta$ represents the model parameter to be optimized. 

By contrast, for the $t$-th task with training data $\mathcal{D}_t$ in CL, given a trained model with the initialized parameter $\theta_{t-1}$, the conventional model finetuning can be achieved by minimizing a classification loss function as: 
\begin{equation}
    \mathcal{L}_{cls}^t = \mathbb{E}_{(x, y) \sim \mathcal{D}_t}[\ell_{ce}(f(x; \theta_{t}), y)].
    \label{eq_loss}
\end{equation}
However, when the previous data is not accessible, simply updating the model with \eqn~\ref{eq_loss} usually leads to a drastic performance drop on previous datasets $\{\mathcal{D}_1, \cdots, \mathcal{D}_{t-1}\}$, \ie catastrophic forgetting.

\subsection{Experience Replay (ER)}
ER~\cite{ICLR2019_Riemer} is a classical rehearsal-based strategy that alleviates forgetting by replaying a subset of previously seen data stored in a permanent memory buffer $\mathcal{M}$. By jointly training the model on current training data $\mathcal{D}_t$ and previous data stored in $\mathcal{M}$, the training loss is computed as:
\begin{equation}
    \mathcal{L} = \mathcal{L}_{cls}^t + \mathcal{L}_{cls}^m,
\end{equation}
where $\mathcal{L}_{cls}^m$ is a classification loss function on the samples stored in the memory buffer and it is used to preserve the acquired knowledge as:
\begin{equation}
    \mathcal{L}_{cls}^m = \mathbb{E}_{(x,y) \sim \mathcal{M}}[\ell_{ce}(f(x; \theta_t), y].
\end{equation}
Theoretically, the larger the memory buffer, the more knowledge would be preserved. For the extreme case, when all the previous data is stored in $\mathcal{M}$, it is equivalent to the joint training in one task. Unfortunately, due to privacy or storage concerns, the previous data might be unavailable in practice, storing a subset of previous data in $\mathcal{M}$ reduces the difficulty of CL.

To leverage the limited memory buffer effectively, ER adopts a reservoir sampling method~\cite{TOMS1985_Vitter} to update the buffer over time. In this way, when we randomly select $|M|$ samples from the input stream, each sample still has the same probability $\frac{|M|}{|S|}$ of being stored in the memory buffer without knowing the length of the input data stream $|S|$ in advance. Although ER alleviates the forgetting issue effectively and outperforms many elaborately designed CL approaches~\cite{NIPS2020_Buzzega}, its performance may drastically drop when few samples are available~\cite{ICCV2021_Verwimp}.

\begin{figure*}[t]
\centering
\includegraphics[width=0.7\textwidth]{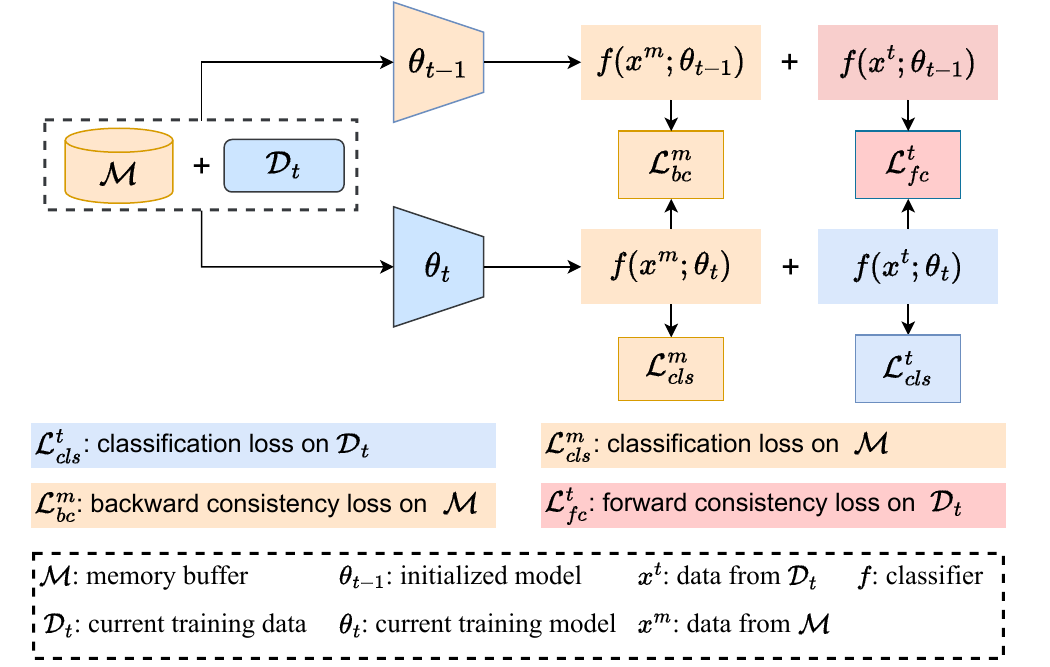}
\caption{Illustration of our SER method. Compared to conventional experience replay methods, the main differences can be summarized as follows. (1) Experience Replay (ER)~\cite{ICLR2019_Riemer}: $\mathcal{L}_{cls}^t + \mathcal{L}_{cls}^m$. (2)  Dark Experience Replay++ (DER++)~\cite{NIPS2020_Buzzega}: $\mathcal{L}_{cls}^t + \mathcal{L}_{cls}^m + \mathcal{L}_{bc}^m$. (3) Strong Experience Replay (SER): $\mathcal{L}_{cls}^t + \mathcal{L}_{cls}^m + \mathcal{L}_{bc}^m + \mathcal{L}_{fc}^m$.}
\label{fig_ser}
\end{figure*}

\subsection{Strong Experience Replay (SER)} 
To further mitigate the forgetting issue, various ER extensions~\cite{NIPS2020_Buzzega, CVPR2022_Wang, ICLR2022_AraniSZ, ECCV2022_Wang} are developed to preserve learned knowledge when learning new tasks. However, these methods may overfit the memory buffer when a few samples are available. In this work, we mix the rehearsal-based strategy with model regularization and we extend ER by incorporating two consistency losses, which helps the CL model consistently evolve during training. Ideally, if an updated model still has the same outputs as its original ones when learning new tasks, it can be considered that the previously learned knowledge is not forgotten. However, this goal cannot be achieved in practice because of disjoint class labels in class-incremental scenarios. Instead, we aim to obtain approximate output logits between the new model and the old one. 


Our method consists of a backward consistency loss and a forward consistency loss. Specifically, the backward consistency $\mathcal{L}_{bc}^m$ is used to distill the model's past experiences on the data stored in the memory buffer. On the other hand, the forward consistency $\mathcal{L}_{fc}^t$ is used to distill the model's future experiences on the current training data, as the new data is from the future for the old model. An overview of the proposed SER method is illustrated in \fig~\ref{fig_ser}. Unlike the backward consistency used in DER++~\cite{NIPS2020_Buzzega}, forward consistency can leverage all training data and it can propagate consistent predictions to new tasks over time. Details of these two consistency losses are as follows.

{\bf Backward consistency.}
For the $t$-th task, given a trained model with initialized parameter $\theta_{t-1}$, we measure the backward consistency between two models on the memory buffer $\mathcal{M}$ with a Mean Square Error (MSE) loss:
\begin{equation}
     \mathcal{L}_{bc}^m = \mathbb{E}_{x \sim \mathcal{M}}[\|f(x; \theta_t) - f(x; \theta_{t-1})\|^2].
     \label{eqn_b}
\end{equation}
As discussed in DER++~\cite{NIPS2020_Buzzega}, the optimization of \eqn~\ref{eqn_b} is equivalent to minimizing the KL divergence loss in knowledge distillation~\cite{Arxiv2015_Hinton}. Instead of using smoothing softmax responses, directly using the logits with an MSE loss can avoid information loss occurring in prediction spaces due to the squashing function (\eg softmax).  

In addition, to reduce the computational cost, we store the logits of the network outputs when updating the memory buffer~\cite{NIPS2020_Buzzega}. Let $z=f(x; \theta)$ denote the output logits of past experience on sample $x$, by replacing $f(x; \theta_{t-1})$ with $z$, then \eqn~\ref{eqn_b} can be rewritten as:
\begin{equation}
     \mathcal{L}_{bc}^m = \mathbb{E}_{x \sim \mathcal{M}}[\|f(x; \theta_t) - z\|^2].
     \label{eqn_bc}
\end{equation}
To clearly illustrate the core idea of the proposed method, we directly use $f(x^m; \theta_{t-1})$ in \fig~\ref{fig_ser}, instead of $z$ stored in the memory buffer. Based on the classification loss on the samples stored in the memory buffer and backward consistency loss, the updated model will mimic its training trajectory and its output logits will approximate its original ones, reducing the forgetting issue.

{\bf Forward consistency.} 
Intuitively, the previous data stored in the memory buffer $\mathcal{M}$ only contains a small subset of $\{\mathcal{D}_1, \cdots, \mathcal{D}_{t-1}\}$. When a few previous samples are available, using a backward consistency alone cannot properly make the outputs of the new model approximate its original ones. As a result, preserving the learned knowledge on a limited memory buffer suffers from the risk of overfitting the $M$, leading to the forgetting problem.  

To address this issue, we design an additional consistency loss to distill future experiences on $\mathcal{D}_t$. In particular, we first compute the output logits of the old model on $\mathcal{D}_t$, and then employ an MSE loss as in \eqn~\ref{eqn_bc} to measure the new consistency as:
\begin{equation}
     \mathcal{L}_{fc}^t = \mathbb{E}_{x \sim \mathcal{D}_t}[\|f(x; \theta_t) - f(x; \theta_{t-1})\|^2].
\end{equation}
Since data in $\mathcal{D}_t$ is from a future task for the old model, we simply refer to $\mathcal{L}_{fc}^t$ as a ``forward consistency'' loss. 

As illustrated in \fig~\ref{fig_consistency}, the backward consistency retains a subset of past experiences stored in $\mathcal{M}$. In contrast, the forward consistency can propagate the future experiences mimicked on $\mathcal{D}_t$ to new tasks over time. Therefore, forward consistency can effectively leverage the complete training data. Moreover, during the model training, the parameters of the old mode are frozen, seeking consistent predictions on the current training data enhances the updated model has a larger overlap with the old model, improving the model generalization on old tasks and further preserving the acquired knowledge.

{\bf Total training loss.} 
Based on the above analysis, the proposed SER method combines two classification losses and two consistency losses during model training. Formally, the total loss used in SER is represented as:  
\begin{equation}
    \mathcal{L} = \mathcal{L}_{cls}^t + \mathcal{L}_{cls}^m + \alpha \mathcal{L}_{bc}^m + \beta \mathcal{L}_{fc}^t
    \label{eqn_ser}
\end{equation}
where $\alpha$ and $\beta$ are the balancing parameters. 

The core challenge in CL is the stability-plasticity dilemma. For the $t$-th task, $\mathcal{L}_{cls}^t$ focuses on the plasticity by finetuning the model on the new dataset $\mathcal{D}_t$; and the other losses focus on the stability by memorizing the experiences of the old model. Specifically, $\mathcal{L}_{cls}^m$ and $\mathcal{L}_{bc}^m$ preserve past experiences on the memory buffer $\mathcal{M}$. $\mathcal{L}_{fc}^t$ retains the future experiences on the current training data $\mathcal{D}_t$. Therefore, compared to the classical ER method, the two consistency losses used in SER effectively improve the stability of the model, and they prevent model changes in prediction spaces and further reduce forgetting.

\begin{algorithm*}[t!]
\caption{Strong Experience Replay.}
\label{alg}
\hspace*{\algorithmicindent} \textbf{Input: }{initialized model parameter $\theta_{t-1}$, dataset $\mathcal{D}_t$, memory buffer $\mathcal{M}$, balancing factors $\alpha$ and $\beta$.} \\
\hspace*{\algorithmicindent} \textbf{Output: }{optimized parameter $\theta_t$, updated memory buffer $\mathcal{M}$.} 
\begin{algorithmic}[1]
\For{all tasks $t \in \{1, 2, \cdots, T\}$}, 
    \For{$(x, y) \in \mathcal{D}_t$},                                \algorithmiccomment{draw current samples}              
        \State $(x', y', z) \gets sample(\mathcal{M})$,              \algorithmiccomment{draw previous samples}
        \State $x^t \gets augment(x)$,                               \algorithmiccomment{data augmentation}
        \State $x^m \gets augment(x')$,                              \algorithmiccomment{data augmentation}
        \State $\mathcal{L}_{cls}^t \gets \ell_{ce}(f(x^t, \theta_t), y)$,   \algorithmiccomment{classification loss on $\mathcal{D}_t$}
        \If{$t ==1$}
            \State $\mathcal{L} \gets \mathcal{L}_{cls}^t$
        \Else
            \State $\mathcal{L}_{cls}^m \gets \ell_{ce}(f(x^m, \theta_t), y')$,  \algorithmiccomment{classification loss on $\mathcal{M}$}
            \State $\mathcal{L}_{bc}^m \gets \|f(x^m; \theta_t) - z\|^2$,          \algorithmiccomment{backward consistency loss on $\mathcal{M}$}            
            \State $\mathcal{L}_{fc}^t \gets \|f(x^t; \theta_t) - f(x^t; \theta_{t-1})\|^2$,  \algorithmiccomment{forward consistency loss on $\mathcal{D}_t$}   
            \State $\mathcal{L} \gets \mathcal{L}_{cls}^t + \mathcal{L}_{cls}^m + \alpha \mathcal{L}_{bc}^m + \beta \mathcal{L}_{fc}^t$,          \algorithmiccomment{final loss for training}              
        \EndIf
        \State $\theta_t$  $\gets$ minimize~$\mathcal{L}$ with SGD,  \algorithmiccomment{model optimization} 
        \State $\mathcal{M} \gets (\mathcal{M}, (x, y, f(x; \theta_t)))$,  \algorithmiccomment{memory update with reservoir sampling}  
    \EndFor
\State $\theta_{t-1} \gets \theta_{t}$, \algorithmiccomment{update previous model parameter} 
\EndFor
\end{algorithmic}
\end{algorithm*}

{\bf Simple Implementation.} 
The proposed SER algorithm is described in \alg~\ref{alg}, where ``aug'' denotes the data augmentation. During model training, we randomly sample one batch of samples from the memory buffer and the augmented data is used for classification and backward consistency. After each training batch, a reservoir sampling method is employed to update the memory buffer. To reduce the computational cost, we store the original training data with its output logits in the memory buffer. It can be seen that SER does not need any complex computation, which makes it easy to implement. Besides, notice that we do not use the consistency losses for the first task, as there is no previous knowledge to be preserved.

\section{Experiments}
We evaluate the proposed method in all three CL settings: Class Incremental Learning ({\bf Class-IL}), Task Incremental Learning ({\bf Task-IL}), and Domain Incremental Learning ({\bf Domain-IL}). For Class-IL, the model needs to incrementally learn new classes in each subsequent task and the task identity is unknown at inference. For Task-IL, the task identity is given during testing. By selecting corresponding classifiers for inference, Task-IL is easier than Class-IL. For Domain-IL, the classes remain the same in each task, but the data distributions are different. 

In our experiments, we strictly follow the experimental settings in DER++~\cite{NIPS2020_Buzzega} and LVT~\cite{CVPR2022_Wang_vit}. To evaluate the performance of our CL method, we do not use any task identity to select the task-specific knowledge at training, even including the Task-IL. Moreover, the network architecture used in our method is fixed for all tasks, thus respecting the constant memory constraint.

\begin{table*}[t]
\centering
\caption{Classification results on CIFAR100 benchmark dataset with a different number of tasks, averaged across 5 runs.}
\resizebox{0.95\textwidth}{!}{
\begin{tabular}{clcccccc}
\toprule
\multirow{2}{*}{\begin{tabular}[c]{@{}l@{}}Buffer  \end{tabular}} & \multirow{2}{*}{Method} & \multicolumn{2}{c}{5 tasks}  & \multicolumn{2}{c}{10 tasks} & \multicolumn{2}{c}{20 tasks}    \\ 
                               &   & Class-IL    & Task-IL      & Class-IL     & Task-IL      & Class-IL & Task-IL \\ \midrule
	               
\multirow{2}{*}{-}     & Joint & 70.21 $\pm$ 0.15 & 85.25 $\pm$ 0.29 & 70.21 $\pm$ 0.15 & 91.24 $\pm$ 0.27 & 71.25 $\pm$ 0.22 & 94.02 $\pm$ 0.33 \\
                       & SGD   & 17.27 $\pm$ 0.14 & 42.24 $\pm$ 0.33 & 8.62 $\pm$ 0.09 & 34.40 $\pm$ 0.53 & 4.73 $\pm$ 0.06 & 40.83 $\pm$ 0.46   \\  \midrule   

\multirow{3}{*}{-}     & oEWC~\cite{ICML2018_Schwarz} & 16.92 $\pm$ 0.28 & 31.51 $\pm$ 1.02 & 8.11 $\pm$ 0.47 & 23.21 $\pm$ 0.49  & 4.44 $\pm$ 0.17 & 26.48 $\pm$ 2.07 \\ 
                       & SI~\cite{ICML2017_Zenke}     & 17.60 $\pm$ 0.09 & {\bf 43.64} $\pm$ 1.11 & 9.39 $\pm$ 0.61 & {\bf 29.32} $\pm$ 2.03  & 4.47 $\pm$ 0.07 & 32.53 $\pm$ 2.70 \\ 
                       & LwF~\cite{TPAMI2017_Li}      & {\bf 18.16} $\pm$ 0.18 & 30.61 $\pm$ 1.49 & {\bf 9.41} $\pm$ 0.06 & 28.69 $\pm$ 0.34  & {\bf 4.82} $\pm$ 0.06 & {\bf 39.38} $\pm$ 1.10 \\   \midrule
	               
\multirow{12}{*}{200}    & ER~\cite{ICLR2019_Riemer}  & 21.94 $\pm$ 0.83 & 62.41 $\pm$ 0.93 & 14.23 $\pm$ 0.12 & 62.57 $\pm$ 0.68 & 9.90 $\pm$ 1.67 & 70.82 $\pm$ 0.74       \\
		         & GEM~\cite{NIPS2017_Lopez} & 19.73 $\pm$ 0.34 & 57.13 $\pm$ 0.94 & 13.20 $\pm$ 0.21 & 62.96 $\pm$ 0.67 & 8.29 $\pm$ 0.18 & 66.28 $\pm$ 1.49      \\
			 & A-GEM~\cite{ICLR2019_chaudhry} & 17.97 $\pm$ 0.26 & 53.55 $\pm$ 1.13 & 9.44 $\pm$ 0.29 & 55.04 $\pm$ 0.87 & 4.88 $\pm$ 0.09 & 41.30 $\pm$ 0.56      \\
			 & iCaRL~\cite{CVPR2017_Rebuffi} & 30.12 $\pm$ 2.45 & 55.70 $\pm$ 1.87 & 22.38 $\pm$ 2.79 & 60.81 $\pm$ 2.48 & 12.62 $\pm$ 1.43 & 62.17 $\pm$ 1.93   \\
			 & FDR~\cite{ICLR2019_Benjamin} & 22.84 $\pm$ 1.49 & 63.75 $\pm$ 0.49 & 14.85 $\pm$ 2.76 & 65.88 $\pm$ 0.60 & 6.70 $\pm$ 0.79 & 59.13 $\pm$ 0.73      \\
			 & GSS~\cite{NIPS2019_Aljundi} & 19.44 $\pm$ 2.83 & 56.11 $\pm$ 1.50 & 11.84 $\pm$ 1.46 & 56.24 $\pm$ 0.98 & 6.42 $\pm$ 1.24 & 51.64 $\pm$ 2.89   \\
			 & HAL~\cite{AAAI2021_Chaudhry} & 13.21 $\pm$ 1.24 & 35.61 $\pm$ 2.95 & 9.67 $\pm$ 1.67 & 37.49 $\pm$ 2.16 & 5.67 $\pm$ 0.91 & 53.06 $\pm$ 2.87     \\
			 & DER++~\cite{NIPS2020_Buzzega} & 27.46 $\pm$ 1.16 & 62.55 $\pm$ 2.31 & 21.76 $\pm$ 0.78 & 63.54 $\pm$ 0.77 & 15.16 $\pm$ 1.53 & 71.28 $\pm$ 0.91    \\
			 & ERT~\cite{ICPR2021_Buzzega} & 21.61 $\pm$ 0.87 & 54.75 $\pm$ 1.32 & 12.91 $\pm$ 1.46 & 58.49 $\pm$ 3.12 & 10.14 $\pm$ 1.96 & 62.90 $\pm$ 2.72 \\   
			 & RM~\cite{CVPR2021_Bang} & 32.23 $\pm$ 1.09 & 62.05 $\pm$ 0.62 & 22.71 $\pm$ 0.93 & 66.28 $\pm$ 0.60 & 15.15 $\pm$ 2.14 & 68.21 $\pm$ 0.43 \\
			 & LVT~\cite{CVPR2022_Wang_vit} & 39.68 $\pm$ 1.36 & 66.92 $\pm$ 0.40 & {\bf 35.41} $\pm$ 1.28 & 72.80 $\pm$ 0.49 &  20.63 $\pm$ 1.14 & 73.41 $\pm$ 0.67  \\      
    	& {\bf SER (Ours)} & {\bf 47.96} $\pm$ 1.03 & {\bf 77.12} $\pm$ 0.34 & 35.29 $\pm$ 1.88 & {\bf 79.85} $\pm$ 0.63 & {\bf 24.35} $\pm$ 1.71 & {\bf 78.86} $\pm$ 1.69  \\   \midrule                        
\multirow{12}{*}{500}  & ER~\cite{ICLR2019_Riemer}  & 27.97 $\pm$ 0.33 & 68.21 $\pm$ 0.29 & 21.54 $\pm$ 0.29 & 74.97 $\pm$ 0.41 & 15.36 $\pm$ 1.15 & 74.97 $\pm$ 1.44    \\ 
	               & GEM~\cite{NIPS2017_Lopez} & 25.44 $\pm$ 0.72 & 67.49 $\pm$ 0.91 & 18.48 $\pm$ 1.34 & 72.68 $\pm$ 0.46 & 12.58 $\pm$ 2.15 & 78.24 $\pm$ 0.61   \\ 
		           & A-GEM~\cite{ICLR2019_chaudhry}  & 18.75 $\pm$ 0.51 & 58.70 $\pm$ 1.49 & 9.72 $\pm$ 0.22 & 58.23 $\pm$ 0.64 & 5.97 $\pm$ 1.13 & 59.12 $\pm$ 1.57         \\ 
	               & iCaRL~\cite{CVPR2017_Rebuffi} & 35.95 $\pm$ 2.16 & 64.40 $\pm$ 1.59 & 30.25 $\pm$ 1.86 & 71.02 $\pm$ 2.54 & 20.05 $\pm$ 1.33 & 72.26 $\pm$ 1.47 \\ 
		           & FDR~\cite{ICLR2019_Benjamin}   & 29.99 $\pm$ 2.23 & 69.11 $\pm$ 0.59 & 22.81 $\pm$ 2.81 & 74.22 $\pm$ 0.72 & 13.10 $\pm$ 3.34 & 73.22 $\pm$ 0.83   \\ 
	               & GSS~\cite{NIPS2019_Aljundi}   & 22.08 $\pm$ 3.51 & 61.77 $\pm$ 1.52 & 13.72 $\pm$ 2.64 & 56.32 $\pm$ 1.84 & 7.49 $\pm$ 4.78 & 57.42 $\pm$ 1.61    \\
	               & HAL~\cite{AAAI2021_Chaudhry}   & 16.74 $\pm$ 3.51 & 39.70 $\pm$ 2.53 & 11.12 $\pm$ 3.80 & 41.75 $\pm$ 2.17 & 9.71 $\pm$ 2.91 & 55.60 $\pm$ 1.83    \\ 	                
                   & DER++~\cite{NIPS2020_Buzzega} & 38.39 $\pm$ 1.57 & 70.74 $\pm$ 0.56 & 36.15 $\pm$ 1.10 & 73.31 $\pm$ 0.78 & 21.65 $\pm$ 1.44 & 76.55 $\pm$ 0.87 \\ 
	               & ERT~\cite{ICPR2021_Buzzega}   & 28.82 $\pm$ 1.83 & 62.85 $\pm$ 0.28 & 23.00 $\pm$ 0.58 & 68.26 $\pm$ 0.83 & 18.42 $\pm$ 1.92 & 73.50 $\pm$ 0.82   \\ 
	               & RM~\cite{CVPR2021_Bang}  & 39.47 $\pm$ 1.26 & 69.27 $\pm$ 0.41 & 32.52 $\pm$ 1.53 & 73.51 $\pm$ 0.89 & 23.09 $\pm$ 1.72 & 75.06 $\pm$ 0.75    \\ 
	               & LVT~\cite{CVPR2022_Wang_vit}  & 44.73 $\pm$ 1.19 & 71.54 $\pm$ 0.93 & 43.51 $\pm$ 1.06 & 76.78 $\pm$ 0.71 & 26.75 $\pm$ 1.29 & 78.15 $\pm$ 0.42    \\                  
	               & {\bf SER (Ours)}  & {\bf 52.23} $\pm$ 0.29 & {\bf 78.94} $\pm$ 0.23 & {\bf 45.17} $\pm$ 1.02 & {\bf 83.73} $\pm$ 0.75 & {\bf 34.14} $\pm$ 0.53 & {\bf 83.69} $\pm$ 1.70    \\                   
 \bottomrule                               
                               
\end{tabular}
}
\label{tbl_cifar100}
\end{table*}                               

\subsection{Experimental Setup}
{\bf Datasets.} The proposed method is evaluated on 5 image classification benchmark datasets. The {\bf CIFAR100} dataset contains 100 classes and each class has 500 images for training and 100 images for testing. {\bf CIFAR10} dataset has 10 classes, each consisting of 5000 training images and 1000 test images. The image size of CIFAR10 and CIFAR100 is $32 \times 32$. {\bf TinyImageNet} has 200 classes and it includes 100,000 training images and 10,000 test images in total, and the image size is $64 \times 64$. Both {\bf Permuted MNIST}~\cite{PANS2017_kirkpatrick} and {\bf Rotated MNIST}~\cite{NIPS2017_Lopez} are built upon the MNIST dataset~\cite{MNIST1998}, which has 60,000 training images and 10,000 test images, and the image size is $28 \times 28$. Permuted MNIST applies a random permutation to the pixels and Rotated MNIST rotates the digits by a random angle in the interval $[0, \pi)$. In our experiments, CIFAR10, CIFAR100, and TinyImageNet are used to evaluate the performance of Class-IL and Task-IL. Permuted MNIST and Rotated MNIST are used for Domain-IL.

{\bf Baselines.} We compare the proposed SER method with three rehearsal-free methods, including oEWC~\cite{ICML2018_Schwarz},  SI~\cite{ICML2017_Zenke}, and LwF~\cite{TPAMI2017_Li}. We also compare SER with multiple rehearsal-based methods: iCaRL~\cite{CVPR2017_Rebuffi}, ER~\cite{ICLR2019_Riemer}, GEM~\cite{NIPS2017_Lopez}, A-GEM~\cite{ICLR2019_chaudhry},  FDR~\cite{ICLR2019_Benjamin}, GSS~\cite{NIPS2019_Aljundi}, HAL~\cite{AAAI2021_Chaudhry}, DER++~\cite{NIPS2020_Buzzega}, Co2L~\cite{ICCV2021_Cha}, ERT~\cite{ICPR2021_Buzzega}, RM~\cite{CVPR2021_Bang}, LVT~\cite{CVPR2022_Wang_vit}, CLS-ER~\cite{ICLR2022_AraniSZ}, and TAMiL~\cite{ICLR2023_Bhat}. Notice that CLS-ER~\cite{ICLR2022_AraniSZ} updates three models (a plastic model, a stable model, and the current training model) during training. In contrast, SER only updates the current training model (the old model is frozen). To show the effectiveness of CL methods, we further provide two baselines without CL techniques: SGD (a lower bound) and Joint (an upper bound). The results of baselines are obtained from the published papers or source codes. Besides, the best parameter configurations of baselines are used for comparison.

{\bf Metrics.} We evaluate the continual learning methods with two metrics: final average accuracy and average forgetting. Let $a_{T,t}$ denote the testing accuracy on $t$-th task when the model is trained on $T$-th task, and the final average accuracy $A_t$ on all $T$ tasks is computed as:
\begin{equation}
Accuracy = \frac{1}{T}\sum_{t=1}^T a_{T,t}. 
\end{equation}
Besides, the average forgetting on $T$ tasks is defined as: 
\begin{equation}
Forgetting =\frac{1}{T-1}\sum_{t=1}^{T-1} max_{i \in \{1, \cdots, T-1}\}(a_{i,t} - a_{T,t}). 
\end{equation}

\subsection{Implementation Details}
{\bf Architectures.} We employ the same network architectures used in DER++~\cite{NIPS2020_Buzzega}. For CIFAR10, CIFAR100, and Tiny ImageNet, we adopt a modified ResNet18 without pretraining. For the MNIST dataset, we adopt a fully connected network with two hidden layers each one comprising 100 ReLU units. 

{\bf Augmentation.} For fairness, we use the same data augmentation as in DER++~\cite{NIPS2020_Buzzega}, which applies random crops and horizontal flips to both stream and buffer examples. It is worth mentioning that small data transformations enforce implicit consistency constraints~\cite{NIPS2020_Buzzega}.

{\bf Training.} Following the training strategy in DER++~\cite{NIPS2020_Buzzega}, we employ a Stochastic Gradient Descent (SGD) optimizer in all experiments. For the CIFAR10 dataset, we train the model with 20 epochs, and CIFAR100 with 50 epochs, respectively. For the Tiny ImageNet dataset, we increase the number of epochs to 100. Besides, we only use one epoch for MNIST variants. The learning rate and batch size are the same as in DER++~\cite{NIPS2020_Buzzega}. In addition, we employ a MultiStepLR to decay the learning rate by parameter 0.1. The milestones are as follows. CIFAR10: [15]; CIFAR100: [35, 45]; TinyImageNet: [70, 90].

Due to the dynamic learning scenarios in continual learning, it is very difficult to adjust the hyper-parameter automatically. Therefore, we follow the previous methods~\cite{NIPS2020_Buzzega, ICLR2022_AraniSZ, ICLR2023_Bhat} that tune the balancing parameters for each dataset. In our experiments, the balancing parameters are as CIFAR10 ($\alpha=0.2$, $\beta=0.2$), CIFAR100 ($\alpha=0.5$, $\beta=0.5$), and TinyImageNet ($\alpha=0.2$, $\beta=1$), P-MNIST and R-MNIST ($\alpha=0.2$, $\beta=0.2$). Other parameter configurations may obtain better performance.

\subsection{Comparison with Previous Methods}
{\bf Evaluation on CIFAR100.} We follow the experimental setting in LVT~\cite{CVPR2022_Wang_vit} to evaluate the proposed method in Class-IL and Task-IL scenarios. \tab~\ref{tbl_cifar100} reports the performance of each CL method on CIFAR100 with 5, 10, and 20 tasks and each task has disjoint class labels. It can be seen that all the rehearsal-free methods fail to reduce forgetting on the CIFAR100 dataset when compared to SGD (a low-bound). Although LwF seeks consistent predictions with knowledge distillation, without the help of previous samples for replaying, the learned model severely drifts over time. In contrast, by storing a small number of previously seen samples in a memory buffer, rehearsal-based methods can effectively preserve the learned knowledge. 

\begin{figure}[t!]
\centering
\begin{subfigure}{.23\textwidth}
  \centering
  \includegraphics[width=\textwidth]{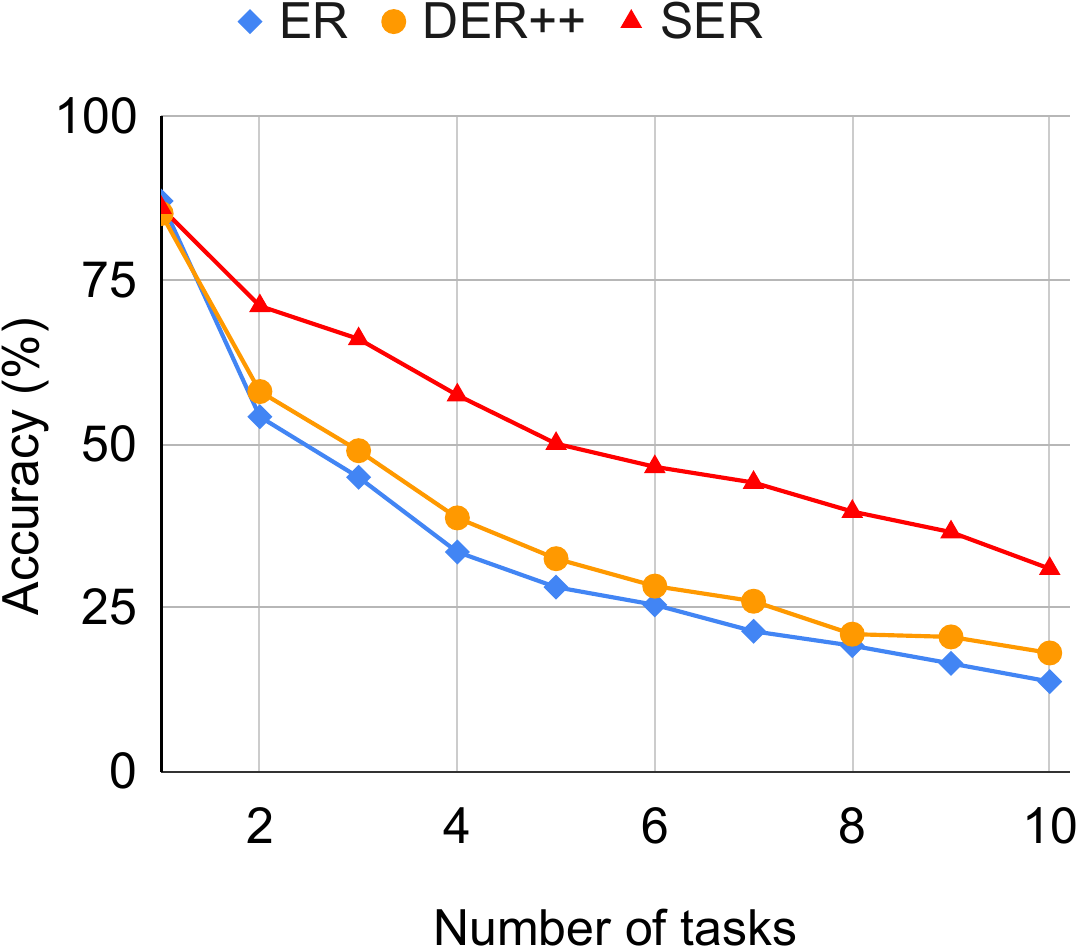}  
  \caption{Class-IL}
\end{subfigure}
\hfill
\begin{subfigure}{.23\textwidth}
  \centering
  \includegraphics[width=\textwidth]{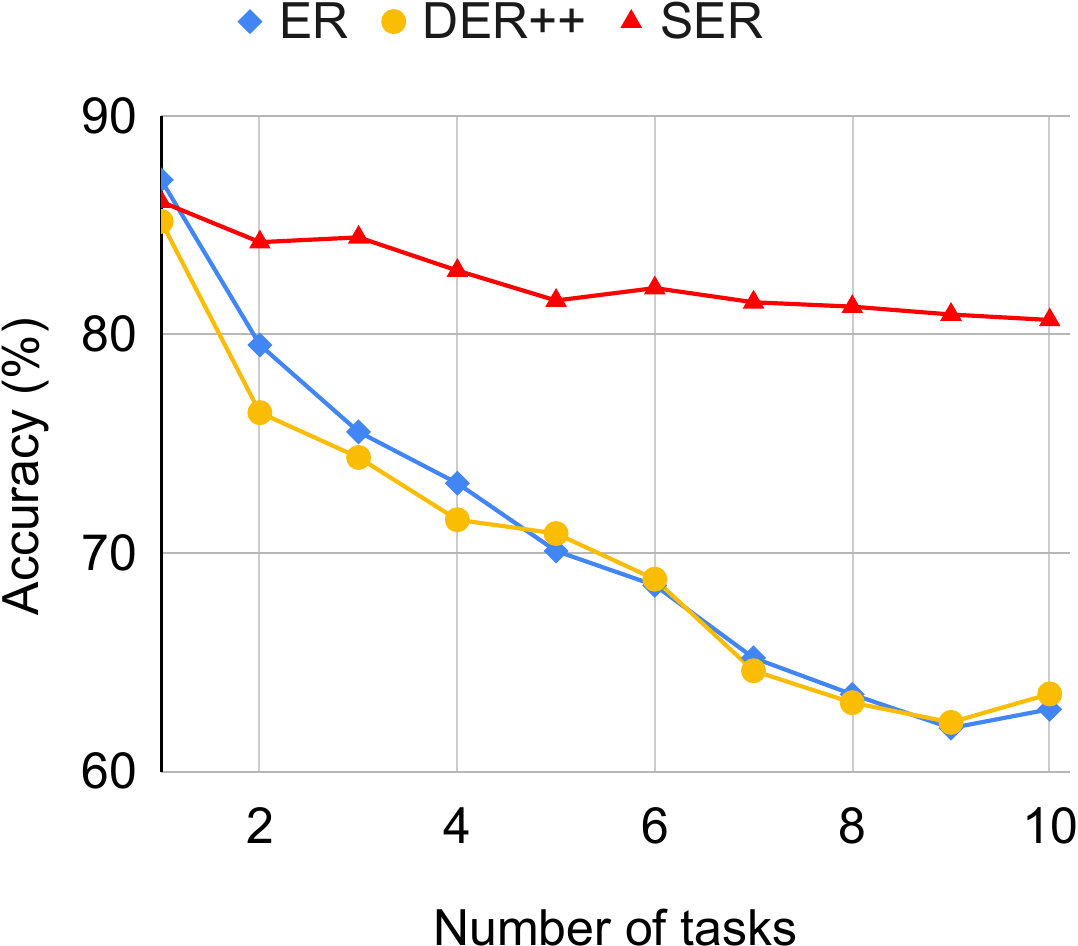}  
  \caption{Task-IL}
\end{subfigure}
\caption{Average accuracy when incrementally learning on the CIFAR100 dataset with 10 tasks (memory size is 200).}
\label{fig_cifar100_10}
\end{figure}

\begin{figure}[t]
\centering
\begin{subfigure}{.23\textwidth}
  \centering
  \includegraphics[width=\textwidth]{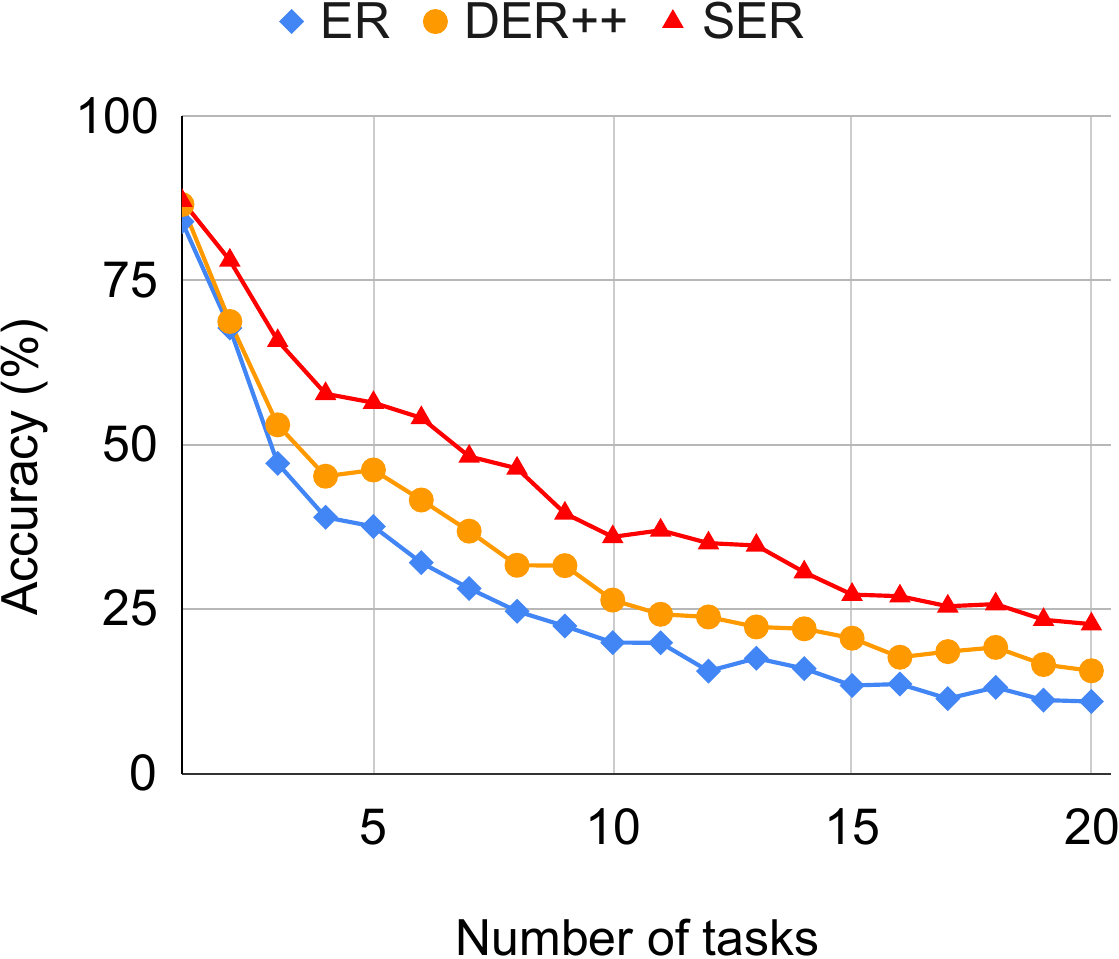}  
  \caption{Class-IL}
\end{subfigure}
\hfill
\begin{subfigure}{.23\textwidth}
  \centering
  \includegraphics[width=\textwidth]{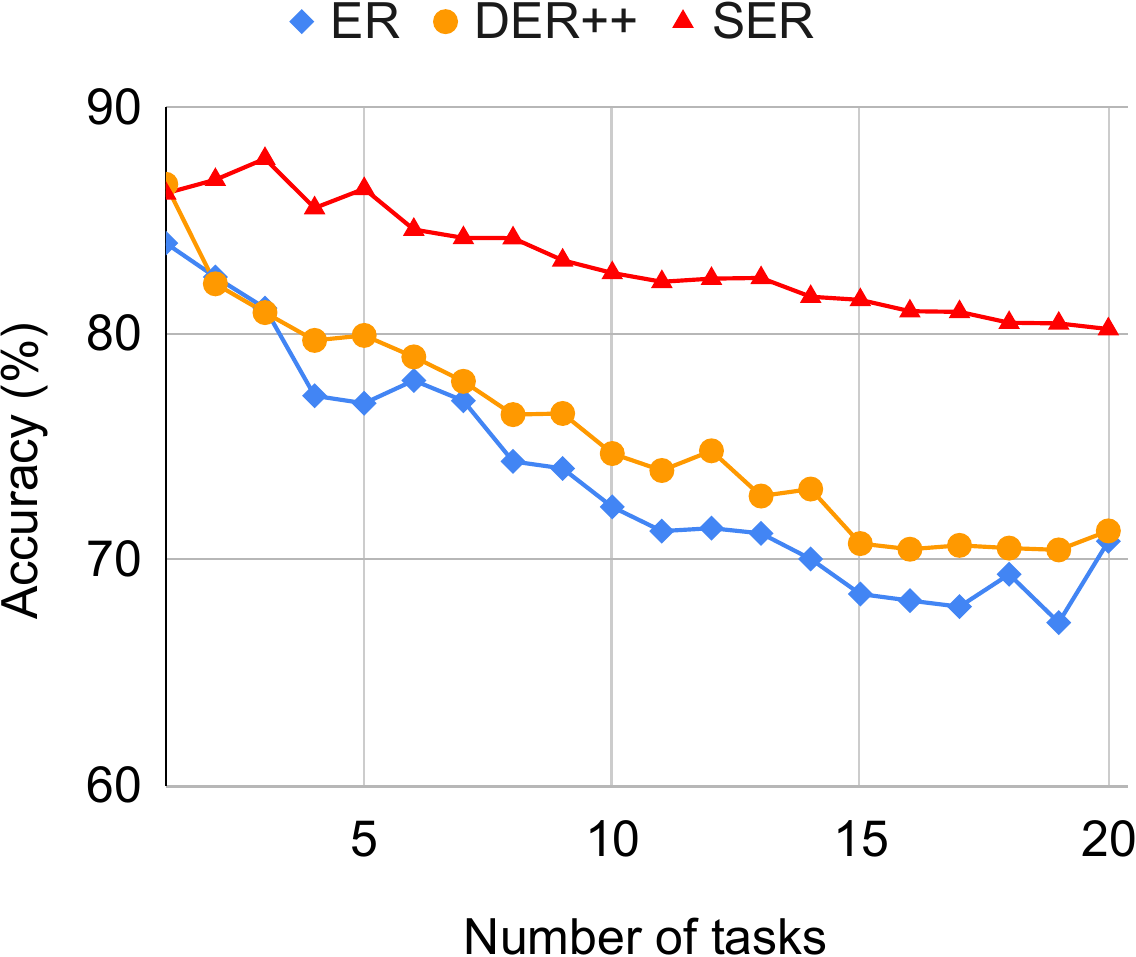}  
  \caption{Task-IL}
\end{subfigure}
\caption{Average accuracy when incrementally learning on the CIFAR100 dataset with 20 tasks (memory size is 200).}
\label{fig_cifar100_20}
\end{figure}

Among the rehearsal-based methods, SER outperforms DER++ by a large margin on all tasks. For example, given a memory buffer with 200 samples for Class-IL, SER outperforms DER++ by 20.5\% on 5 tasks and 9.2\% on 20 tasks. Besides, for Task-IL with 10 tasks, SER achieves a 16.3\% accuracy improvement. \fig~\ref{fig_cifar100_10} and \fig~\ref{fig_cifar100_20} plot the average accuracy of  ER, DER++, and SER on CIFAR100 when learning each task. It can be seen that SER substantially improves the accuracy in both Class-IL and Task-IL for each learning stage. By adding a forward consistency loss, DER++ surpasses ER by a noticeable margin in both Class-IL and Task-IL. Moreover, with the help of a forward consistency loss, SER substantially outperforms DER++ on CIFAR100 with different lengths of tasks.

\begin{table*}[t]
\centering
\caption{Classification results on standard CL benchmarks, averaged across 5 runs. Due to the task identity requirement, LwF, iCaRL, and TAMiL cannot be applied in the Domain-IL setting. Besides, the training time of GEM, HAL, and GSS are intractable on TinyImageNet. Co2L uses a more complex network architecture on the variants of MNIST datasets, we do not report its accuracy for a fair comparison. }
\resizebox{0.95\textwidth}{!}{
\begin{tabular}{clcccccc}
\toprule
\multirow{2}{*}{\begin{tabular}[c]{@{}l@{}}Buffer \end{tabular}} & \multirow{2}{*}{Method} & \multicolumn{2}{c}{CIFAR10}  & \multicolumn{2}{c}{TinyImageNet} & P-MNIST   & R-MNIST   \\ 
                               &   & Class-IL    & Task-IL      & Class-IL     & Task-IL      & Domain-IL & Domain-IL \\ \midrule

\multirow{2}{*}{-}   & Joint             & 92.20 $\pm$ 0.15 & 98.31 $\pm$ 0.12 & 59.99 $\pm$ 0.19 & 82.04 $\pm$ 0.10 & 94.33 $\pm$ 0.17 & 95.76 $\pm$ 0.04  \\
                     & SGD               & 19.62 $\pm$ 0.05 & 61.02 $\pm$ 3.33 & 7.92  $\pm$ 0.26 & 18.31 $\pm$ 0.68 & 40.70 $\pm$ 2.33 & 67.66 $\pm$ 8.53  \\   \midrule
\multirow{3}{*}{-}   & oEWC~\cite{ICML2018_Schwarz}   & 19.49 $\pm$ 0.12 & {\bf 68.29} $\pm$ 3.92 & 7.58  $\pm$ 0.10 & 19.20 $\pm$ 0.31 & {\bf 75.79} $\pm$ 2.25 & {\bf 77.35} $\pm$ 0.04  \\
                     & SI~\cite{ICML2017_Zenke}       & 19.48 $\pm$ 0.17 & 68.05 $\pm$ 5.91 & 6.58  $\pm$ 0.31 & {\bf 36.32} $\pm$ 0.13 & 65.86 $\pm$ 1.57 & 71.91 $\pm$ 8.53  \\  
                     & LwF~\cite{TPAMI2017_Li}        & {\bf 19.61} $\pm$ 0.05 & 63.29 $\pm$ 2.35 & {\bf 8.46}  $\pm$ 0.22 & 15.85 $\pm$ 0.58 & -                & -                 \\  \midrule                 
\multirow{12}{*}{200} & ER~\cite{ICLR2019_Riemer}      & 44.79 $\pm$ 1.86 & 91.19 $\pm$ 0.94 &  8.49 $\pm$ 0.16 & 38.17 $\pm$ 2.00 & 72.37 $\pm$ 0.87 & 85.01 $\pm$ 1.90  \\
                     & GEM~\cite{NIPS2017_Lopez}      & 25.54 $\pm$ 0.76 & 90.04 $\pm$ 0.94 &  -               &     -            & 66.93 $\pm$ 1.25 & 80.80 $\pm$ 1.15  \\
                     & A-GEM~\cite{ICLR2019_chaudhry} & 20.04 $\pm$ 0.34 & 83.88 $\pm$ 1.49 &  8.07 $\pm$ 0.08 & 22.77 $\pm$ 0.03 & 66.42 $\pm$ 4.00 & 81.91 $\pm$ 0.76  \\                     
                     & iCaRL~\cite{CVPR2017_Rebuffi}  & 49.02 $\pm$ 3.20 & 88.99 $\pm$ 2.13 &  7.53 $\pm$ 0.79 & 28.19 $\pm$ 1.47 & -                & -                 \\        
                     & FDR~\cite{ICLR2019_Benjamin}   & 30.91 $\pm$ 2.74 & 91.01 $\pm$ 0.68 &  8.70 $\pm$ 0.19 & 40.36 $\pm$ 0.68 & 74.77 $\pm$ 0.83 & 85.22 $\pm$ 3.55  \\        
                     & GSS~\cite{NIPS2019_Aljundi}    & 39.07 $\pm$ 5.59 & 88.80 $\pm$ 2.89 &   -              &  -               & 63.72 $\pm$ 0.70 & 79.50 $\pm$ 0.41  \\        
                     & HAL~\cite{AAAI2021_Chaudhry}   & 32.36 $\pm$ 2.70 & 82.51 $\pm$ 3.20 &   -              &  -               & 74.15 $\pm$ 1.65 & 84.02 $\pm$ 0.98  \\        
                     & DER++~\cite{NIPS2020_Buzzega}  & 64.88 $\pm$ 1.17 & 91.92 $\pm$ 0.60 & 10.96 $\pm$ 1.17 & 40.87 $\pm$ 1.16 &  83.58 $\pm$ 0.59 & 90.43 $\pm$ 1.87  \\
                     & Co2L~\cite{ICCV2021_Cha}       & 65.57 $\pm$ 1.37 & 93.43 $\pm$ 0.78 & 13.88 $\pm$ 0.40 & 42.37 $\pm$ 0.74 & - & -  \\   
                     & CLS-ER~\cite{ICLR2022_AraniSZ} & 66.19 $\pm$ 0.75 & 93.59 $\pm$ 0.87 & 23.47 $\pm$ 0.80 & 58.41 $\pm$ 1.72 & {\bf 84.63} $\pm$ 0.40 & {\bf 92.26} $\pm$ 0.18  \\   	
                     & TAMiL~\cite{ICLR2023_Bhat}	&	68.84 $\pm$ 1.18	&	94.28 $\pm$ 0.31	&	20.46 $\pm$ 0.40	&	55.44 $\pm$ 0.52	 & - & - \\  
                     & {\bf SER (Ours)}    & {\bf 69.86} $\pm$ 1.28 & {\bf 94.51} $\pm$ 0.61 & {\bf 28.50} $\pm$ 1.21 & {\bf 70.12} $\pm$ 0.67 & 82.76 $\pm$ 0.62 & 91.78 $\pm$ 0.81  \\                        \midrule        
\multirow{12}{*}{500} & ER~\cite{ICLR2019_Riemer}      & 57.74 $\pm$ 0.27 & 93.61 $\pm$ 0.27 &  9.99 $\pm$ 0.29 & 48.64 $\pm$ 0.46 & 80.60 $\pm$ 0.86 & 88.91 $\pm$ 1.44  \\
                     & GEM~\cite{NIPS2017_Lopez}      & 26.20 $\pm$ 1.26 & 92.16 $\pm$ 0.69 &  -               &     -            & 76.88 $\pm$ 0.52 & 81.15 $\pm$ 1.98  \\
                     & A-GEM~\cite{ICLR2019_chaudhry} & 22.67 $\pm$ 0.57 & 89.48 $\pm$ 1.45 &  8.06 $\pm$ 0.04 & 25.33 $\pm$ 0.49 & 67.56 $\pm$ 1.28 & 80.31 $\pm$ 6.29  \\                     
                     & iCaRL~\cite{CVPR2017_Rebuffi}  & 47.55 $\pm$ 3.95 & 88.22 $\pm$ 2.62 & 9.38  $\pm$ 1.53 & 31.55 $\pm$ 3.27 & -                & -                 \\        
                     & FDR~\cite{ICLR2019_Benjamin}   & 28.71 $\pm$ 3.23 & 93.29 $\pm$ 0.59 & 10.54 $\pm$ 0.21 & 49.88 $\pm$ 0.71 & 83.18 $\pm$ 0.53 & 89.67 $\pm$ 1.63  \\        
                     & GSS~\cite{NIPS2019_Aljundi}    & 49.73 $\pm$ 4.78 & 91.02 $\pm$ 1.57 &   -              &  -               & 76.00 $\pm$ 0.87 & 81.58 $\pm$ 0.58  \\        
                     & HAL~\cite{AAAI2021_Chaudhry}   & 41.79 $\pm$ 4.46 & 84.54 $\pm$ 2.36 &   -              &  -               & 80.13 $\pm$ 0.49 & 85.00 $\pm$ 0.96  \\        
                     & DER++~\cite{NIPS2020_Buzzega}  & 72.70 $\pm$ 1.36 & 93.88 $\pm$ 0.50 & 19.38 $\pm$ 1.41 & 51.91 $\pm$ 0.68 & 88.21 $\pm$ 0.39 & 92.77 $\pm$ 1.05  \\
                     & Co2L~\cite{ICCV2021_Cha}       & 74.26 $\pm$ 0.77 & {\bf 95.90} $\pm$ 0.26 & 20.12 $\pm$ 0.42 & 53.04 $\pm$ 0.69 &  -  & -   \\     
                     & CLS-ER~\cite{ICLR2022_AraniSZ} & 75.22 $\pm$ 0.71 & 94.35 $\pm$ 0.38 & 31.03 $\pm$ 0.56 & 61.57 $\pm$ 0.63 & 88.30 $\pm$ 0.14 & {\bf 94.06} $\pm$ 0.07  \\                    
                     & TAMiL~\cite{ICLR2023_Bhat}	&	74.45 $\pm$ 0.27	&	94.61 $\pm$ 0.19	&	28.48 $\pm$ 1.50	&	65.19 $\pm$ 0.82 & - & - \\
                     & {\bf SER (Ours)}  & {\bf 75.67} $\pm$ 0.90 & 95.67 $\pm$ 0.56 & {\bf 33.20} $\pm$ 0.84 & {\bf 71.53} $\pm$ 1.14 & {\bf 88.58} $\pm$ 0.51 & 92.14 $\pm$ 0.63  \\              
\bottomrule    
\end{tabular}
}
\label{tbl_all}
\end{table*}

Even compared to the most recent LVT method in Class-IL, SER surpasses it by a large margin of 8.3\% on 5 tasks and 3.0\% on 20 tasks when 200 previous samples are stored. Moreover, SER obtains at least 6\% accuracy improvement in Task-IL. Furthermore, given a larger memory buffer with 500 samples, it can be observed that SER also obtains consistent accuracy improvement. These results demonstrate that SER can effectively preserve preserves the learned knowledge with a strong experience replay. 

{\bf Evaluation on CIFAR10.} \tab~\ref{tbl_all} shows the performance of each CL method on the CIFAR10 dataset (5 tasks). Similar to the results observed in \tab~\ref{tbl_cifar100}, rehearsal-based methods surpass the rehearsal-free methods by a large margin and our SER achieves the best performance among compared algorithms. CIFAR10 is a small dataset that contains 10 classes. When the memory size is 200, the average number of stored samples for each class is 20. In contrast, for the CIFAR100 dataset with 100 classes, there are only 2 samples for each class. Therefore, by storing more samples for each class, the average accuracy on CIFAR10 is much better for all CL approaches. 

Compared to DER++, SER further reduces the forgetting issue and boosts performance with the forward consistency loss. For example, SER outperforms DER++ by 5\% and 3.7\% in Class-IL when the memory size is 200 and 500, respectively. Notice that CLS-ER is a dual memory experience replay architecture, which computes a consistency loss on the data stored in the memory buffer and maintains short-term and long-term semantic memories with episodic memory. By contrast, our SER incorporates a forward consistency loss on current training data only. Compared to CLS-ER on CIFAR10, SER improves the accuracy by 3.7\% when the memory size is 200. Furthermore, although SER does not employ the task identities during model training, it slightly outperforms the most recent method TAMiL by 1\%.

{\bf Evaluation on TinyImageNet.} In \tab~\ref{tbl_all}, the TinyImageNet data is divided into 10 tasks and each task has 20 classes. TinyImageNet is more challenging than CIFAR10 and CIFAR100 for CL when a few previous samples are available. As the results reported in \tab~\ref{tbl_all}, given a limited memory buffer with a buffer size of 200, which indicates that there is only 1 sample for each class on average. Compared to the low-bound baseline SGD (7.92\% in Class-IL), all rehearsal-free methods and most rehearsal-based methods cannot effectively reduce forgetting with a noticeable margin. Because there are a few samples for each class only, it is very difficult to retain the acquired knowledge from limited past experiences. Therefore, these rehearsal-based methods fail to reduce forgetting. 

By mimicking the future experiences of the old model on current training data, SER can distill additional knowledge and further reduce forgetting. From \tab~\ref{tbl_all}, we can observe that SER substantially improves the performance on TinyImageNet when compared to most previous CL methods. In particular, the average accuracy of SER is much better than DER++ in Class-IL, \ie 28.50\% \vs 10.96\%. For Task-IL, SER significantly outperforms DER++ with an absolute improvement of 29.25\%. Compared to another recent method Co2L, SER also obtains 14.6\% accuracy improvement in Class-IL and more than 27.7\% in Task-IL. Moreover, SER surpasses CLS-ER and TAMiL by a noticeable margin of 5.03\% and 8.04\% in Class-IL, respectively. These results suggest that SER can effectively preserve the learned knowledge when few samples are available for experience replay.

\begin{table}[t]
\centering
\caption{Average forgetting of ER and its expansions on CIFAR10 (5 tasks) and CIFAR100 (20 tasks), lower is better. Memory size is 200 for both two datasets.}
\begin{tabular}{ccccc} \toprule
\multirow{2}{*}{Method} & \multicolumn{2}{c}{CIFAR10} & \multicolumn{2}{c}{CIFAR100} \\
                        & Class-IL      & Task-IL     & Class-IL      & Task-IL      \\ \midrule
ER~\cite{ICLR2019_Riemer}       &   54.12   &   5.72      &  83.54    &  23.72       \\
DER++~\cite{NIPS2020_Buzzega}   &   39.84   &   8.25      &  71.02    &  20.64        \\
{\bf SER (Ours)}                & {\bf 20.25} & {\bf 2.52} & {\bf 60.61} & {\bf 13.61}  \\  \bottomrule        
\end{tabular}
\label{tbl_forget}
\end{table}

\begin{figure*}[!t]
\centering
\includegraphics[width=1.0\textwidth]{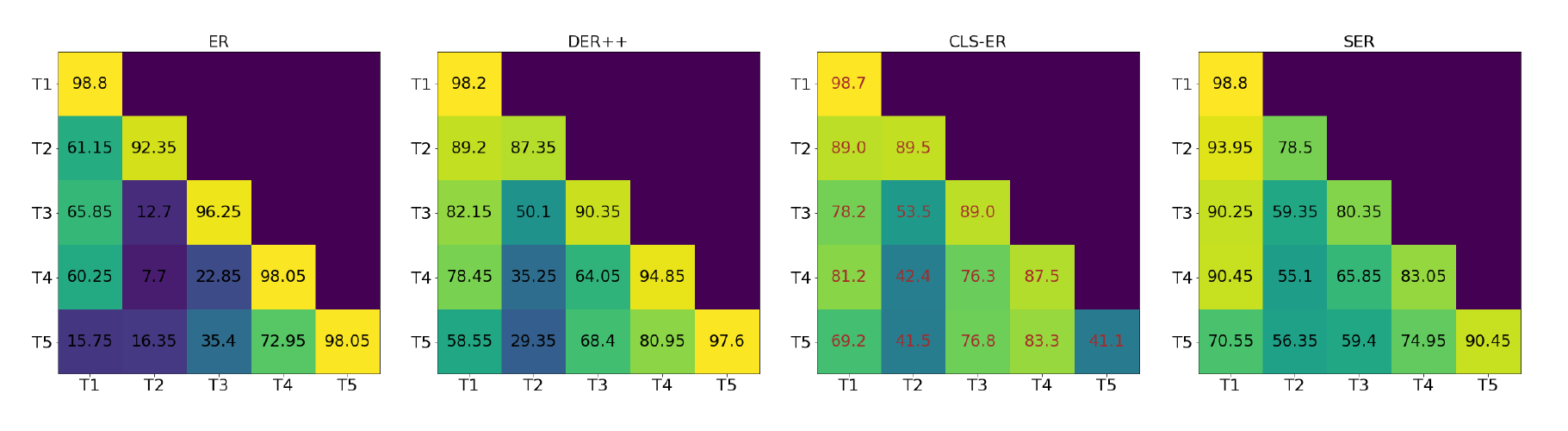}
\caption{Accuracy of different CL methods after sequentially learning each task on CIFAR10 dataset in Class-IL scenarios. The models are
evaluated at the end of each task (y-axis) to evaluate how the task performances (x-axis) are affected as training progress. The size of the memory buffer is 200. }
\label{fig_matrix}
\end{figure*}

{\bf Evaluation on P-MNIST and R-MNIST.} To evaluate the performance of the proposed method in Domain-IL, we follow the experimental settings in DER++\cite{NIPS2020_Buzzega}, which adopts 20 tasks for Permuted MNIST (P-MNIST) and Rotated MNIST (R-MNIST). As the average accuracy reported in \tab~\ref{tbl_all}, based on a memory buffer with 500 samples, most rehearsal-based methods outperform the rehearsal-free methods (except A-GEM) on both two datasets. However, by reducing the memory size to 200, the performance of ER is slightly worse than oEWC on P-MNIST. In contrast, DER++ and SER are also effective in Domain-IL by seeking consistent predictions. Besides, CLS-ER achieves the best performance on P-MNIST and R-MNIST and it is comparable to SER and DER++ on both two datasets. 

Based on these observations, it can be concluded that the proposed forward consistency is more effective in Class-IL and Task-IL than Domain-IL. The reason is that the category spaces in Class-IL and Task-IL are disjoint while it is the same in Domain-IL. Therefore, seeking consistency with the current training data may overwrite the previous knowledge and it does not further mitigate forgetting.

{\bf Average Forgetting.}
Besides average accuracy, forgetting is another important metric to measure the performance of a CL method. \tab~\ref{tbl_forget} shows the average forgetting of ER, DER++, and our SER. It can be observed that SER achieves the lowest forgetting in all the comparison experiments, including both Class-IL and Task-IL on two datasets. These results indicate that SER effectively reduces the forgetting issue, verifying the effectiveness of our proposed forward consistency loss.

\begin{table*}[t]
\centering
\caption{Ablation study on each loss of SER with average accuracy. The memory size is 200. CIFAR10 consists of 5 tasks and CIFAR100 consists of 20 tasks. According to the losses used in previous methods, some loss combinations can be represented as (1) SGD: $\mathcal{L}_{cls}^t$; (2) ER: $\mathcal{L}_{cls}^t+\mathcal{L}_{cls}^m$; (3) DER++: $\mathcal{L}_{cls}^t+\mathcal{L}_{cls}^m+\mathcal{L}_{bc}^m$.}
\begin{tabular}{cccc|cccc} \toprule
\multicolumn{4}{c}{Loss} & \multicolumn{2}{c}{CIFAR10} & \multicolumn{2}{c}{CIFAR100} \\ 
$\mathcal{L}_{cls}^t$  & $\mathcal{L}_{cls}^m$ & $\mathcal{L}_{bc}^m$  & $\mathcal{L}_{fc}^t$  & Class-IL      & Task-IL      & Class-IL       & Task-IL       \\ \midrule
\Checkmark &            &            &            & 19.26  & 61.02  & 4.73  &  40.83 \\
\Checkmark & \Checkmark &            &            & 44.79  & 91.19  & 9.90  &  70.82  \\
\Checkmark & \Checkmark & \Checkmark &            & 64.88  & 91.92  & 15.16 &  71.28  \\
\Checkmark & \Checkmark &            & \Checkmark & 69.04  & {\bf 95.79}  & 19.50 &  79.82 \\
\Checkmark & \Checkmark & \Checkmark & \Checkmark & {\bf 69.86}  & 95.51   & {\bf 24.35}  & {\bf 79.86} \\ \bottomrule
\end{tabular}
\label{tbl_ablation}
\end{table*}

\subsection{Ablation Study}
To further analyze the effectiveness of each loss in our method, we report the ablation study results in \tab~\ref{tbl_ablation}. Without storing any previous samples, simply finetuning the model with $\mathcal{L}_{cls}^t$ on current training data nearly forgets all previous knowledge in Class-IL, which is only 19.26\% on CIFAR10 and 4.73\% on CIFAR100. By replaying a small number of previous samples, ER effectively reduces the forgetting issue on both two datasets with $\mathcal{L}_{cls}^t + \mathcal{L}_{cls}^m$. Besides, based on the backward consistency loss $\mathcal{L}_{bc}^m$, DER++ further reduces forgetting by seeking consistency predictions on the data stored in the memory buffer. Furthermore, fusing two classification losses and two consistency losses, \ie $\mathcal{L}_{cls}^t+\mathcal{L}_{cls}^m+\mathcal{L}_{bc}^m +\mathcal{L}_{fc}^t$, SER achieves the best performance on all the comparison experiments.

It is worth mentioning that simply extending ER with a forward consistency loss $\mathcal{L}_{fc}^t$ alone also surpasses DER++. For example, from the results on CIFAR10 reported in \tab~\ref{tbl_ablation}, the average accuracy is improved from 64.88\% to 69.04\% on Class-IL and 91.92\% to 95.79\% on Task-IL. These observations indicate that seeking consistent predictions on current training data can also help to preserve the acquired knowledge. Moreover, the forward consistency can be computed on the complete training data and it can propagate the consistent distributions to new tasks over time. Therefore, the forward consistency loss demonstrates better performance when a small memory buffer is given. 

\subsection{Stability-Plasticity Dilemma}
To better understand how well different CL methods strike a balance between stability and plasticity, we report the performance of several CL methods after sequentially learning 5 tasks on the CIFAR10 dataset in \fig~\ref{fig_matrix}. In terms of average accuracy, DER++, CLS-ER, and SER outperform ER with the help of consistency loss, showing the effectiveness of stability. On the other hand, ER achieves better accuracy on the current training data, and it outperforms other methods on plasticity. Based on the proposed forward consistency, SER improves the model generalization on previous tasks and it maintains a better balance. Compared to DER++ and CLS-ER, SER preserves more knowledge of previous tasks and it also achieves good performance on the recent task. Therefore, SER outperforms the compared methods with the metric of average accuracy, verifying its effectiveness.

\subsection{Computational Cost Analysis}
Based on the same backbone, the computational cost of different CL methods relies on the efficiency of model training. Since we adopt the less or the same training epochs as the baseline methods~\cite{NIPS2020_Buzzega, ICLR2022_AraniSZ}, we mainly analyze the computational cost by comparing the model design. As illustrated in \alg~\ref{alg}, besides the basic model optimization on the current training data, we draw one batch of samples from the memory buffer for both the experience replay and backward consistency at the same time. In contrast, DER++~\cite{NIPS2020_Buzzega} draws two batches of samples from the memory buffer, where one batch of samples is for the experience replay while the other is for backward consistency. Therefore, although we incorporate a forward consistency loss by computing the output logits of the old model on the current training data, SER does not increase the computational cost when compared to DER++. By contrast, CLS-ER~\cite{ICLR2022_AraniSZ} adopts a working model for experience replay, and it additionally computes the consistency with the memory buffer by a stable model and a plastic model. Compared to the computational cost of CLS-ER, the proposed SER method is more efficient.

\subsection{Discussion}
The proposed SER method mixes the experience replay and model regularization in CL. Compared to the previous rehearsal-based methods, we additionally design a forward consistency loss to improve the model generalization on previous tasks. Since forward consistency can leverage more data to reduce drastic model changes during training and propagate the past logit distributions to new tasks over time, the forgetting issue is effectively reduced. Moreover, the proposed method is easy to implement, and its computational cost is not expensive when compared to the state-of-the-art methods, \eg DER++ and CLS-ER. Therefore, the proposed SER method is simple and effective. We hope it can improve the incremental learning ability of other deep models in the future.

\section{Conclusion}
We propose a Strong Experience Replay (SER) that distills the learned knowledge from both past and future experiences. Besides memorizing the ground truth labels of previous samples stored in the memory buffer, SER also seeks consistent predictions between the new model and the old one on the data from both the memory buffer and the input stream. By imposing a backward and a forward consistency loss into model training, SER effectively reduces the forgetting issue when a few samples are available. Extensive experiments demonstrate that SER outperforms state-of-the-art methods by a large margin. Moreover, ablation study results on two datasets show the effectiveness of the proposed forward consistency.

\bmhead{Acknowledgments}
This research is supported by the National Natural Science Foundation of China under Grant 62002188, and the Shandong Excellent Young Scientists Fund Program (Overseas) 2023HWYQ-114.

\bmhead{Data Availability}
All data sets used in this article are publicly available. Data sharing does not apply to this article as no new data sets were generated during the current study.

\bibliographystyle{plain}
\bibliography{sn-bibliography}

\end{document}